# Potential Auto-driving Threat: Universal Rain-removal Attack

Jinchegn Hu[a,1], Jihao Li[a,1], Zhuoran Hou[a], Jingjing Jiang[a], Cunjia Liu[a] and Yuanjian Zhang[a]*

[a] Department of Aeronautical and Automotive Engineering, Loughborough University, Leicestershire, LE11 3TU, United Kingdom

**Abstract**

The problem of robustness in adverse weather conditions is considered a significant challenge for computer vision algorithms in the applicants of autonomous driving. Image rain removal algorithms are a general solution to this problem. They find a deep connection between raindrops/rain-streaks and images by mining the hidden features and restoring information about the rain-free environment based on the powerful representation capabilities of neural networks. However, previous research has focused on architecture innovations and has yet to consider the vulnerability issues that already exist in neural networks. This research gap hints at a potential security threat geared toward the intelligent perception of autonomous driving in the rain. In this paper, we propose a universal rain-removal attack (URA) on the vulnerability of image rain-removal algorithms by generating a non-additive spatial perturbation that significantly reduces the similarity and image quality of scene restoration. Notably, this perturbation is difficult to recognise by humans and is also the same for different target images. Thus, URA could be considered a critical tool for the vulnerability detection of image rain-removal algorithms. It also could be developed as a real-world artificial intelligence attack method. Experimental results show that URA can reduce the scene repair capability by 39.5% and the image generation quality by 26.4%, targeting the state-of-the-art (SOTA) single-image rain-removal algorithms currently available.
**Keywords:** Autonomous Driving, Rain Removal, Deep Learning, Computer Vision;

## 1. Introduction

As more autonomous driving companies gradually embark on commercial landing projects and more passenger cars and trucks are equipped with assisted and autonomous driving features to actively play a role in real-world scenarios such as online vehicles [1], intercity delivery [2], long-distance [3] and high-speed deliveries [4], more and more people are expressing high expectations for autonomous driving and demonstrating its great commercial potential [5]. While AI has made excellent progress in ADAS and autonomous driving tasks, its robustness and safety in deployment environments are still not widely recognised by society, and research and validation of AI for complex environment awareness has never stopped [6]. Of concern is the heavy reliance on deep learning (DL) models for environment perception in autonomous driving systems. The RGB images captured by on-board cameras are the critical perceptual path for intelligent perception of autonomous driving, but perception capabilities are often affected by rain in the driving environment [7]. As a complex atmospheric process, rainfall during vehicle driving will result in unpredictable visibility degradation. Often, raindrops/streaks close to the optical sensor may obscure or distort the scene content in the image, while rain streaks deeper in the scene often produce large areas of tiny streaks of obscuration or fog etc., which in turn blur the image content [8]. This reduced visibility will create a potential safety threat to AI systems deployed in autonomous driving systems, such as pedestrian trajectory prediction [9], scene understanding [10], target recognition [11] and semantic segmentation [12], increasing the likelihood of safety incidents. Therefore, removing rain streaks and raindrops from environment-aware RGB images to restore scene information, i.e. image rain-removal, has become a key pre-processing tool to enhance AI-aware safety. As a key research topic to ensure AI safety during autonomous driving, many DL-based image rain removal studies have yielded excellent results and attracted significant attention in the fields of computer-vision and pattern recognition [13]. However, related research has shown that adversarial examples initially designed to influence generic DL models can also be used to cause failures in autonomous driving tasks [14]. In related cases, some imperceptible image perturbations are often added to the perceptual image of the camera to reduce the accuracy of the DL model or to increase the functional error rate of other AI systems [15]. Such adversarial threats may also be present in image rain-removal algorithms. From an autonomous driving perspective, any inaccurate results of the image rain-removal algorithm due to perturbations may compromise driving safety. More seriously, malicious perturbation attacks will be difficult to identify in cases where humans do not easily detect perturbations, allowing perpetrators to get away with it. The hazards of imperceptible human perturbations on DL-based AI systems have been well discussed in previous studies. Still, their impact on DL-based image rain-removal algorithms has not been verified and discussed in depth. This research gap would prevent the application of autonomous driving in countries or regions with high average annual precipitation.

---

* Corresponding author. *E-mail address:* y.y.zhang@lboro.ac.uk.
[1]These authors contributed equally to this work.

This paper aims to analyse potential threats to DL-based rain removal algorithms, validate their vulnerability and demonstrate their potential hazards. We propose a threat validation method for DL-based rain-removal algorithms to facilitate this research. The proposed method incorporates a non-additive perturbation generation method based on image spatial-transform. Perturbations constructed based on this spatial-transform rule are applied to the rain-observed image to interfere with the rain-removal results, which are analysed in terms of human eye observation, pixel distribution and AI detection.

Two key technical challenges are addressed in implementing this validation method: 1) Non-additive perturbation construction. Unlike adding noise directly to the image pixels, non-additive perturbation is achieved by swapping and fusing pixel values to achieve interference. The proposed method describes the direction and step of pixel changes in rain-observed images by generating a flow field comparable to the image size. 2) Optimisation of perturbation generation for DL-based image rain-removal. The proposed method designs an end-to-end generative model and constructs loss functions based on luminance, contrast and structural differences from the perspective of human visual understanding. The generative model will maximise the loss between the unperturbed and perturbed rain removal results.

To explore the potential threat of such perturbations to intelligent RGB image-based perception systems in autonomous driving tasks, we validated RainCCN [16], a neural network model with advanced performance in the image rain-removal research, as the subject. Its perturbation results were subjected to adequate human eye observation, pixel distribution analysis and computer-vision AI detection. By connecting the experimental results with relating the pixel distribution, we demonstrate the vulnerability of DL-based image rain-removal algorithms. Specifically, the contributions of this paper are as follows:

1. This paper introduces potential security threats for the first time, to our knowledge, to image rain-removal algorithms and gives hypothetical attack scenarios. The experimental results in this paper clarify the potential security threat of image rain-removal algorithms. It is a pain point for AI applications in autonomous driving. The hypothetical attack scenarios are given to elaborately describe the vulnerability detection and exploitation process of image rain-removal algorithms, forming a classic case study for AI-aware security research in autonomous driving.
2. The possible robustness problem of DL-based image rain-removal algorithms is demonstrated. This paper proposes a universal perturbation generation method for the image rain-removal problem, which generates a unique non-additive perturbation through a neural network. The perturbation will distort the pixel distribution of the image rain removal dataset. By analysing the performance difference of the rain removal algorithm beforeand after the perturbation, the robustness problem of the DL-based image rain removal algorithm is clarified.
3. The proposed perturbation generation method provides new evidence of the impact of sensor degradation on computer-vision systems and the safety of autonomous driving. The unique and non-noticeable perturbation corresponds to potential sensor degradation. It can be used for malicious attacks and assessing sensor degradation's impact in autonomous driving applications.

The rest of the paper is organised as follows: Section 2 introduces the studies related to this work. Section 3 describes the general process of image DL-based rain-removal algorithms and presents a hypothetical attack scenario to clarifies the potential threats to DL-based image rain-removal algorithms in this attack scenario. Section 4 describes in detail a non-additive perturbation generation method for the image rain-removal algorithm. Section 5 shows the experimental results. Section 6 expounds the limitations of this work and points to future research. Section 7 gives conclusions.

**2. Related Work**

In this section we review related work from the perspectives of the image rain-removal and adversarial attacks in computer vision.

*2.1. Image Rain-removal*

The image rain-removal solutions can be divided into physical attribute-based rain-removal algorithms and deep learning-based rain-removal algorithms. The detection and removal of raindrops from images have become to be a more and more vital research area based on the physical properties of rainwater. Falling raindrops are subjected to many physical conditions and thus deformed, such as surface tension, hydrostatic pressure, ambient illumination and aerodynamic pressure [17]. These irregular perturbations appear as raindrops/rain streaks of different brightness/orientation and contaminate the scene information in the image [18]. Beard and Chuang [19] proposed the equilibrium shape model of raindrops which described a kinematic model of raindrops in images. In addition to this, physical information such as luminance, chromaticity and temporal space have been developed for image raindrop/rainstreaks detection and elimination. However, there is room for improvement in the restoration of scene information in physical property-based image rain-removal methods because no reasonable linkage with scene information has been established [20]. In order to figure out how to use image filters to introduce image scene information in physical property-based rain-removal methods to enhance scene restoration. Xu et al. [21] proposed an enhancement scheme for chromaticity rain-removal based on guided filters. In addition to this sparse coding-based dictionary learning, histograms of oriented gradients-based [22] and a priori-based [23] approaches have been proposed successively to enhance the scene restoration capability of rain removal algorithms. As a data-driven feature self-learning method, Deep learning has a good advantage in learning image scene information. Aiming to remove dirt and water droplets adhering to glass windows or camera lenses, Eigen et al. [24] proposed a convolutional neural network-based (CNN-based) image rain-removal algorithm. However, the method was unable to handle the dense raindrops and dynamic raindrops and produced blurred output. [25] designed a fine-grained generative network to cope with the presence of a large number of raindrops. This work introduces visual attention mechanisms into the design of the generative adversarial network [26, 27]. The generative network focuses on the raindrop region and its surroundings which is used to generate a similar image with the surroundings of the background images and without raindrops at the same time, and the discriminative network was mainly used to evaluate the similarity between the rain-removal images and the clear images. Qian et al. [28] designed a CNN-based approach, DerainNet, to specifically handle individual image rain streak removal, which automatically learns a non-linear mapping function between clean and rain image details from the dataset. The Deep Residual Network (ResNet) extends the structural depth of neural networks with powerful feature learning capabilities [29]. Fu et al. [30] proposed a Deep Detail Network (DDN) to reduce the range of mappings from input to output through the introduction of ResNet. Similarly, Fan et al. [31] proposed a residual-guided feature fusion network (ResGuideNet) that gradually obtains coarse to fine estimates of negative residuals as the network progresses. Zhang and Patel [32] further proposed a density-aware image drainage method using a multi-stream dense network (DID-MDN), which can adapt the different rain density by integrating a residual-aware classifier process. Li et al. [33] proposed a cyclic squeeze and excitation-based contextual aggregation network (CAN). The CAN for individual image rainwater removal, where the SE block not only assigns different values to various rain streak layers but also make the CAN obtain large receptive fields and have a good performance for adapting to the rainwater removal task. The previous research has placed great emphasis on exploring the structure of deep neural networks (DNNs), and they have aimed to improve the network structure to achieve better image recovery.

*2.2. Adversarial sample*

Goodfellow et al. [34] proposed the Fast Gradient Sign Method (FGSM), which uses gradients to generate adversarial samples. The samples contained small perturbations that were imperceptible to humans, but the DNNs produced highly confident incorrect answers to these inputs. They further demonstrated that an attacker could create adversarial inputs for specific labels. Later researchers showed how scaled gradient iterations could be applied to the original input image. Madry et al. [35] proposed an iterative method, the projected gradient descent (PGD) attack, which showed better attack performance than FGSM. The Carlini and Wagner attack (CW) [36] describe how to generate adversarial samples by solving the optimization problem efficiently.

Moosavi-Dezfooli et al. [37] proposed an important work to generate universal adversarial perturbations (UAP) by the neural network. The UAP integrates perturbations learned from each iteration. If the combination fails to mislead the target model, UAP will perform a new perturbation and then project the new perturbation onto the $l2$ norm sphere to ensure that the new perturbations are sufficiently small and satisfies the distance limit. This method will continue to run until the empirical error in the sample set is sufficiently large or the threshold error rate is satisfied.

Hayes and Danezis et al. [38] further developed UAP and they proposed the universal adversarial network (UAN). It is composed of a deconvolution layer, a batch normalization layer with an activation function and several fully connected layers at the top. The UAN includes a distance minimisation term in the objective function and the size of the generated noise is controlled by a scaling factor. Poursaeed et al. [39] used the ResNet-based generator to generate a generic adversarial perturbation (GAP) on the semantic segmentation tasks. Mopuri et al. [40] proposed an adversary generative model called NAG, which proposes an objective function that aims to reduce the confidence in benign predictions and increase confidence in other categories. There is a diversity term is introduced into the objective function to encourage diversity in perturbations.

Fawzi and Frossard [41] suggests that convolutional neural networks are not robust in terms of rotation, translation and expansion, which demonstrates the potential threat of pixel space variation to deep learning models. Xiao et al. [42] further proved this view, they argued that the traditional *ln* constraint might not be an ideal measure of similarity between two images. Therefore, they proposed an optimization method for spatial transformation that is able to generate perceptually realistic adversarial examples with high deception rates by changing the position of pixels rather than adding perturbations directly to a clean image. It manipulates the target image according to a pixel substitution rule called "flow field", which maps each pixel in the image to another. In order to ensure the perturbated image is perceptually close to the clear image, they also minimize local geometric distortion in the objective function. Their experimental results show that this non-additive perturbation (flow field) is more difficult for humans to detect than additive perturbations such as [37] and [38].

In previous research, DNNs have been shown to be potentially vulnerable and can be exploited to trigger erroneous output. The generic and non-additive methods of attack are more concealable and destructive. These studies provide a valuable reference for the methodology to be proposed in this paper and provide the methodological basis and feasibility rationale.

## 3. General DL-based rain-removal process

Deep learning has from the outset focused on mining potential information connections between image pixels, and its powerful feature self-learning capabilities have been demonstrated in a variety of image restoration tasks. In the image rain removal task, DL-based image rain removal methods are able to accurately identify raindrops/streaks and restore key scene elements that are obscured/blurred. This section clarifies the general process of DL-based image rain-removal methods.

*3.1. Image observation in a rainy scene*

Although images captured on rainy days in the real world often contain complex, unpredictable and incomprehensible blocks (sets) of pixels, in the potential consensus reached by the rain-removal algorithm, the observed image of a rainy day can be represented as:

$$O = B + R \tag{1}$$

where $O$ is the observation image of the monocular camera in the rain scene, $B$ represents the image information of the background environment, and $R$ represents the blurred mask with raindrops/streaks. There are three manifestations of rain information in rain scenes:

- When the background environment continues to rain and the image perception element protects the glass is not attached to the raindrops, the obtained perceptual image will be covered by rain streaks:

$$R = R_s \tag{2}$$

where $R_s$ denotes the pixel information of rain streaks in the background.

- In the case of a background scene where rain has stopped and there are raindrops adhering to the glass, the perceptual image obtained will be obscured by the raindrops and the scattering of light caused by the raindrops will further obscure the scene information in the image:

$$R = (1-M) \odot B + R_D - B \tag{3}$$

where $R_D$ denotes the pixel information of raindrops on the glass of the image perception element. $m$ is a blurring mask of the same size as the background image, which takes values between [0, 1].

- In the case of a background scene with continuous rain and the presence of raindrops adhering to the glass, the perceptual image obtained will be blurred by the raindrops while being obscured by the rain streaks:

$$R = (1-M) \odot (B + R_s) + \eta R_D - B \tag{4}$$

where $\eta$ represents the atmospheric scattering constant, and relevant studies have proved that atmospheric light can be simplified to constant terms in image clarity research.

In general, the combined effect of rain tracks and raindrops will minimise visibility and corrupt scene information in images, negatively impacting on the environmental perception and cognitive abilities of humans and AI.

*3.2. DL-based end-to-end rain-removal solution*

Compared to traditional machine learning, which requires complex feature analysis and data processing methods, DL has a powerful feature self-learning capability. It can adopt an end-to-end supervised learning model with complex neural network architecture design and hyperparametric filtering to achieve perceptual capabilities beyond human vision. In neural network-based image rain-removal solutions, end-to-end neural network models are widely used for feature learning and reduction with the help of convolutional and deconvolutional neural networks, with a general objective function of:

$$J_\theta = \underset{\theta}{argmin}\, L(\theta(O), B) \tag{5}$$

where $\theta$ denotes the end-to-end neural network; $\theta(O)$ denotes the rain-removal sample; L denotes the loss function, being used to describe the difference between samples; and $J_\theta$ describes the optimization objective of the neural network. The improvement of the loss function is a key approach to maximize the image scene information reduction capability of deep neural networks, where the introduction of structural similarity of images (SSIM) is considered to be the most suitable objective function design for the needs of the human visual system. SSIM describes the degree of similarity of image samples by comparing local luminance, contrast and structure:

$$\begin{cases} SSIM(x,y) = l(x,y)^\alpha \cdot c(x,y)^\beta \cdot s(x,y)^\gamma \\ l(x,y) = \dfrac{2\mu_x \mu_y + c_1}{\mu_x^2 + \mu_y^2 + c_1} \\ c(x,y) = \dfrac{2\sigma_x \sigma_y + c_2}{\sigma_x^2 + \sigma_y^2 + c_2} \\ s(x,y) = \dfrac{\sigma_{xy} + c_3}{\sigma_x \sigma_y + c_3} \end{cases}, x = \theta(O), y = B \tag{6}$$

where $x$ represents a rain-removal image sample and $y$ represents a clear ambient image sample. $l(x, y)$ is the luminance contrast function; $c(x, y)$ is the contrast comparison function; and $s(x, y)$ is the structure contrast function. Where $\mu_x$ and $\mu_y$ describe the luminance of the sample; $\sigma_x$ and $\sigma_y$ describe the contrast of the sample; $c_1$, $c_2$, $c_3$ are the fractional stability constants. In general, the larger the SSIM value, the better the quality of the scene information reproduction the easier it is to be recognised and understood by humans. Therefore, together with a high-performance gradient optimiser, it is easier to achieve good rain-removal performance by using the negative value of SSIM as the objective function for neural network optimisation:

$$L_\theta(\theta(O), B) = L_1(\theta(O), B) - \lambda SSIM(\theta(O), B) \tag{7}$$

where $L_1$ denotes the mean absolute value error between the rain-removal sample and the background image sample; $\lambda$ denotes the importance factor of SSIM. $L_1$ loss ensures that the neural network optimiser reduces the absolute bias of the pixels, and SSIM ensures that the neural network optimiser attempts to reduce the visual comprehension bias.

*3.3. Rain-removal performance evaluation*

In addition to evaluating sample similarity using SSIM, this paper evaluates image rain-removal performance in terms of both image sharpness and semantic understanding. Peak Signal-to-Noise Ratio (PNSR), as a metric to quantify the reconstruction quality of images and videos affected by lossy compression, is used to express the ratio between the maximum possible power of a signal and the power of corrupted noise that affects the fidelity of its representation:

$$PSNR(\theta(O), B) = 10 \cdot \log_{10}\left(\frac{MAX_B^2}{MSE(\theta(O), B)}\right) \quad (8)$$

where *MSE* means the mean square error; $MAX_B$ indicates the maximum possible pixel value of the image. In general, the higher the value of PSNR the better the quality of the image reconstruction.

*3.4. The life cycle of DL-based rain removal*

As a classic AI system, Figure 1 illustrates the common lifecycle of a DL-based rain removal algorithm. In an image rain-removal task, the data access, storage and delivery will be strictly limited and managed from data acquisition to model evaluation, which is a production line process. However, when the system is deployed the AI system, the DL-based rain-removal algorithm, will access data from external sources. At this point in its lifecycle, the closed loop of secure development is broken and it is formally exposed to external threats.

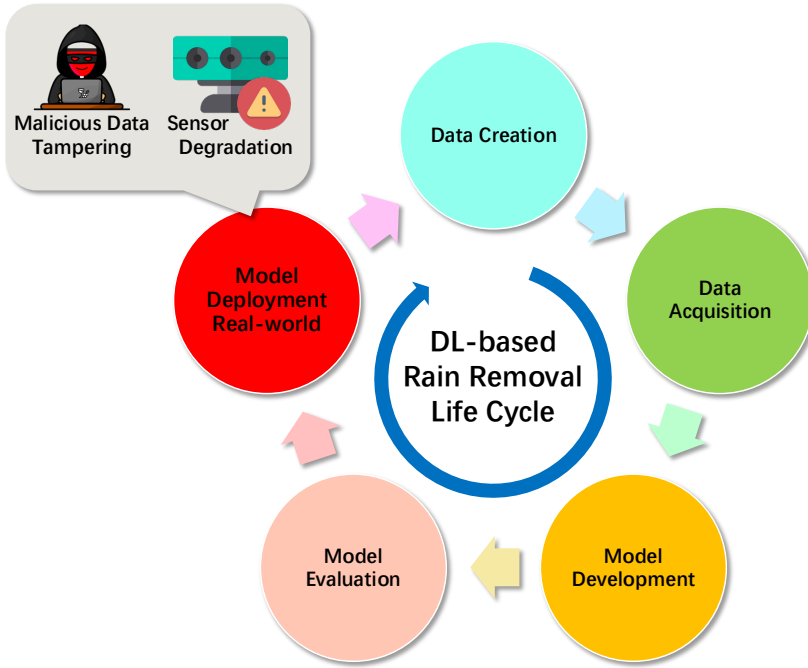

Figure 1. Life cycle of the DL-based rain-removal algorithm.

*3.5. Potential attack scenarios*

By analysing the lifecycle of the DL-based rain-removal algorithm, this paper constructs a potential attack scenario as shown in Figure 2. External malicious adversarial attacks and sensor degradation are common threats to in-vehicle artificial intelligence. In the proposed attack scenario, the normal rain-removal module would generate a high-quality clean background image based on the rain observation image. However, in both malicious attacks and sensor degradation, an unnoticeable noise can be applied while or after the camera captures an image. This noise can affect subsequent intelligent computer-vision-based perception modules and thus potentially increase the risk of accidents in autonomous driving. The ability to identify or exclude the effect of noise on rain removal results is a gap in current research on DL-based rain removal algorithms. Assuming that noise is not readily recognisable by humans and can significantly corrupt image rain-removal results, such noise would significantly increase the difficulty of accident investigation, even if accident investigators are unable to determine the presence of noise or the presence of a potentially malicious attacker:
- The generate perturbations that are difficult for humans to detect.
- The perturbation has universal rain-removal attack (URA) capabilities and has the potential for realistic applications.
- The perturbed rain observation image can significantly degrade the rain removal performance of DL-based images.

Based on this potential attack scenario, this paper identifies the basic objectives of the proposed attack method.

## 4. End-to-end universal non-additive perturbation generation

This section describes an end-to-end attack method for DL-based rain-removal by generating an image space perturbation that cannot be detected by humans in order to degrade the rain-removal performance. The proposed method consists of two main modules: 1) an image spatial change module. It is responsible for updating the pixel values in a specified direction of spatial variation and by a prescribed interpolation method. 2) A deep neural network-based pixel flow field generator. It will generate the pixel flow field for a data-driven image rain-removal algorithm. Its general process is:

$$O_{adv} = ST(0, U_\vartheta) \tag{9}$$

where $\vartheta$ denotes the neural network employed in the proposed method, which maximises the disturbance capacity of the perturbation by learning the dataset features; $U_\vartheta$ denotes the Universal flow field (UFF) output by the neural network, which describes the direction and magnitude of the spatial transformation of the image pixels; and ST denotes the spatial transformation rule of the image, which applies the perturbation to the rain-observed image according to the indication of the UFF.

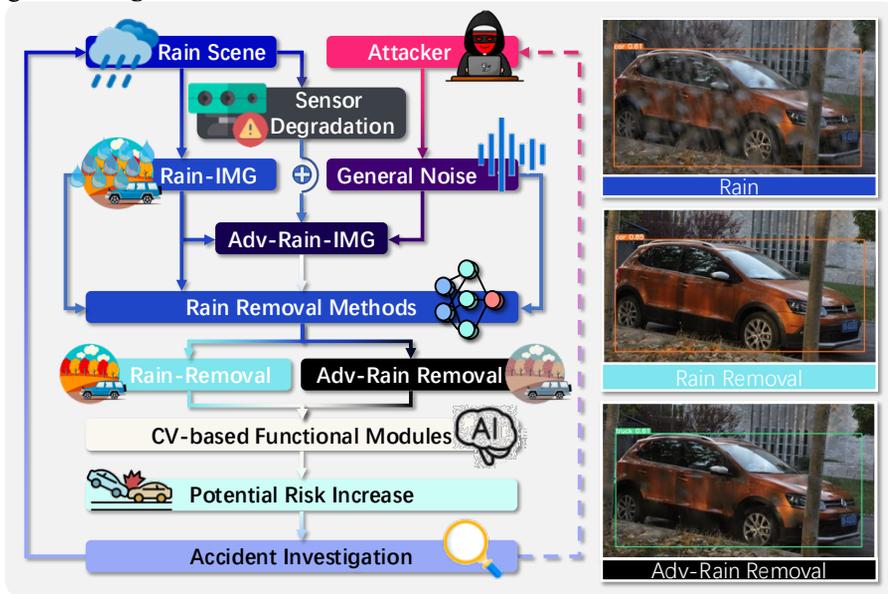

Figure 2. Potential attack vector.

## 4.1. Image Spatial Transformation

Spatially transform (ST) is an image-specific attack method in computer-vision, it is proposed in it proposed by [42]. Unlike applying noise directly to the image, the method generates an example of an adversarial attack by manipulating each pixel value based on a learned flow field. In the method the rain-observed image $O$ is regularised and its input space is $I$ [0, 255] ($c·h·w$), $c$ denotes the image channel, in this paper only RGB images are considered, $h$ is the image height and $w$ is the image length. UFF is a matrix with the same geometry as the input image, as it only considers the horizontal and vertical shift relationship of each pixel without considering the difference of the RGB channels, thus $UFF$ [0, 1] ($2·h·w$). For any pixel of any channel in the image, the UFF describes the following spatial transformation process:

$$(u_i, v_i) = (u_i' + (h-1)\Delta_i^u, v_i' + (w-1)\Delta_i^v), UFF_i = (\Delta_i^u, \Delta_i^v), i \in \{1, 2, \dots, h \times w\} \tag{10}$$

where $i$ denotes the index of any pixel within a channel, and $(u, v)$ denotes the pixel coordinate position. $\Delta$ denotes the step size of the pixel space transformation. To ensure that the spatially transformed image is not severely distorted, the transformed image pixels are updated by means of neighbourhood pixel interpolation, the validity of which is verified by [42, 43], whose calculation process is:

$$p_i^{st} = ST(p_i, UFF_i) = \sum_{q \in N} p_q (1 - |u_i' - u_q|)(1 - |v_i' - v^{(q)}|), p_q \in 0, p_i^{st} \in O_{adv} \tag{11}$$

where $p_i$ is the pixel involved in the spatial transformation; $p_i^{st}$ denotes the target pixel; $N$ denotes the set of pixels at the top left, top right, bottom left and bottom right of the target pixel location.

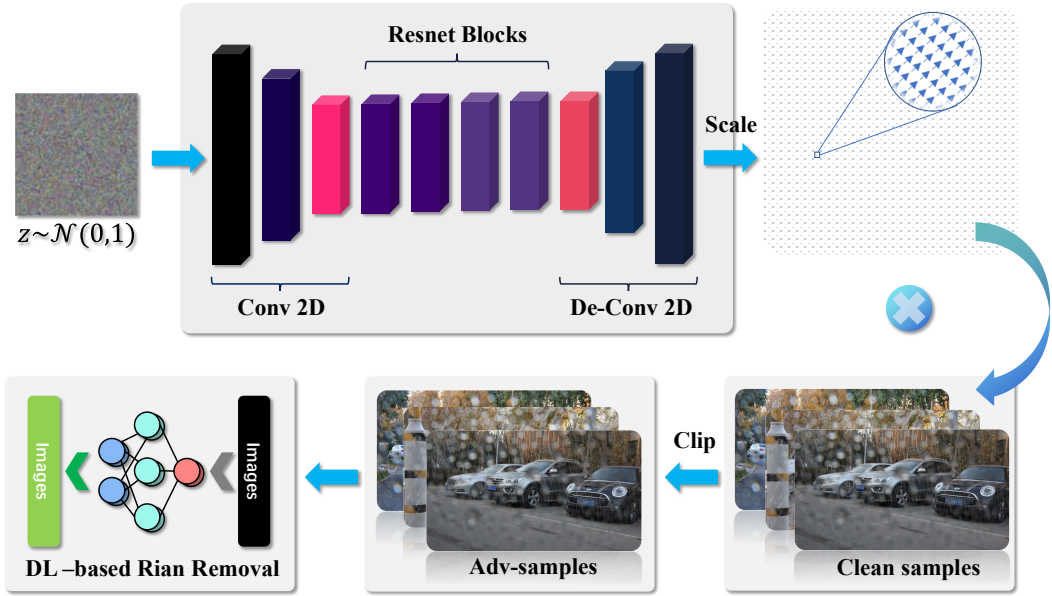

Figure 3. Attack Methodology.

## 4.2. DL-based Flow Filed Generation

Image spatial variations and pixel flow fields guide the direction of offset and update patterns for each pixel value in an image. Manually scripting target-oriented pixel shifts for such large pixel populations (720P images) requires significant practical experience and time costs, and it is difficult to guarantee its generality across different images. This paper uses a flow field generation model based on a DNN to generate UFF for the image rain removal algorithm. As shown in the figure, the proposed neural network includes three modules:
- ○ Data dimension reduction module. A three-layer convolutional neural network can reduce the input high-dimensional Gaussian noise by setting different convolutional kernel sizes and step lengths. The instance normal layer and Relu layer further help the convolutional layer improve model generalization and feature representation.
- ○ Feature learning module. ResNet has been shown to be a high-performance neural network structure for mitigating gradient disappearance or gradient explosion in deep neural networks. Four ResNet-32 models are concatenated in series in the feature learning module used in this paper. The robust feature learning capability of the large-depth ResNet network will capture the mapping of Gaussian noise to the underlying flow field in the hidden state transfer of the neural network.
- ○ Flow field generation module. In contrast to the data dimension reduction module, this module attempts to understand the output of the feature learning module, transforming it into a flow field of a specified size by means of a three-layer deconvolution network. The introduction of the Sigmoid module ensures that any value of the generated flow field is scaled to [0,1].

- Table 1. Pseudocode of the Algorithm 1.

**Algorithm 1:** Universal rain-removal attack.
**Input:** Training set $O$, clear images $B$, total epochs $T$, initial input pattern $z$, adversarial generation model $\vartheta$, target, model $\theta$, and number of mini-batches $M$, learning rate $l$
**Output:** A universal flow field $U_\vartheta$
　for do $t = 1...T$
　　for do $i = 1...M$
　　　$o = O_i, b = B_i, z = \mathcal{N}(0,1)$　　　　▷ Load Data
　　　$U_\vartheta = \vartheta(z)$　　　　　　　　　　　▷ Generate a flow field
　　　$o_{adv} = ST(o)$　　　　　　　　　　　　▷ Perform spatial transformation
　　　$o_{adv} = \text{Clip}(o_{adv}, 0, 255)$　　　　　▷ Clip pixel value
　　　$\vartheta = \vartheta - l\nabla_\vartheta \mathcal{L}_\vartheta(o_{adv}, o, b)$　　　　▷ Gradient update

　end for
end for

The proposed attack method intends to disrupt DL-based image rain-removal capabilities and reduce the understanding of computer-vision-based AI in rain-removal images. This paper introduces SSIM as a tool for image similarity detection that simulates human visual understanding to construct a loss function:

$$\mathcal{L}_\vartheta(O_{adv}, O, B) = \text{SSIM}(\theta(O_{adv}), \theta(0)) + \varphi \text{SSIM}(\theta(O_{adv}), B) \quad (12)$$

where $\vartheta$ is an end-to-end neural network generation model, $\varphi$ is the loss balance coefficient, which implies the extent to which real scene elements are considered in the generation of the UFF; $\theta(O_{adv})$ denotes the rain-removal image of the perturbed sample. The objective function of neural network $\vartheta$:

$$J_\vartheta = \underset{\vartheta}{\text{argmin}} \mathcal{L}_\vartheta(O_{adv}, O, B) \quad (13)$$

The optimization process of its objective function is shown in Table 1.

## 5. Results

This paper develops complete experiments in image rain-removal, attack perturbation generation and AI performance interference. This section describes the hardware and software configurations and hyper-parameter settings that the experiments rely on and analyses the results.

### 5.1. Experimental Setting

A workstation was used to run all experiments in this work, which included two Intell Xeon E5-2620 processors and an Nvidia RTX 3090-24G GPU. All code in this paper is based on Python, where all neural networks are built, trained and deployed based on PyTorch. In order to constrain the experimental cost, this paper takes RainCCN as the attack object and RainDS [10], which contains 1000 photos collected in real, as the data set for training and verification. Under the current computing power conditions, it takes an average of 0.23 seconds to attack 16 images based on the generated universal perturbations, and its attack frequency can reach about 4.23 Hz. The main hyperparameters used in this work are shown in Table 2. RainCCN was trained from scratch as an attack target, using the RainDS training set as input, with 16 images at a time and updated at a learning rate of 0.001. This paper does not deeply study the training process of the DL-based rain-removal algorithm, but only focuses on its final rain-removal performance. After 100 iterations of the training set, RainCCN's learning rate decayed to 0.000001. URA, a deep neural network-based generative model, iterates through 200 epochs in the RainDS training set at a learning rate of 0.01, sampling 100 samples from the training set for each update. The Adam optimizer was used to train this neural network model with an L2 regularization weight of 0.0001 and a momentum gradient calculation constant ($\beta_1$, $\beta_2$) of 0.5,0.9 respectively. The SSIM fractional calculation constants $c_1$, $c_2$ and $c_3$ are 0.0001, 0.0009 and 0.00045 respectively.

Table 2. Hyperparameters

| Algorithm | Item | Description | Value |
|---|---|---|---|
| RainCCN | lr | Learning rate for RainCCN training | $1\times10^{-3}$ |
| RainCCN | Min lr | Minimum learning rate for RainCCN training | $1\times10^{-5}$ |
| RainCCN | Epoch | Number of epochs to train for RainCCN | 100 |
| RainCCN | batch size | Training batch size | 16 |
| URA | lr | Learning rate for UFF training | 0.01 |
| URA | batch size | Training batch size | 100 |
| URA | epoch | Number of epochs to train for UFF | 200 |
| URA | l2reg | Weight factor for l2 regularization | $1\times10^{-3}$ |
| URA | beta1 | $\beta_1$ for the Adam | 0.5 |
| URA | Beta2 | $\beta_2$ for the Adam | 0.9 |
| SSIM | $c_1$ | Fractional calculation constants | $1\times10^{-4}$ |
| SSIM | $c_2$ | Fractional calculation constants | $9\times10^{-4}$ |
| SSIM | $c_3$ | Fractional calculation constants | $4.5\times10^{-4}$ |

In addition to using SSIM and PSNR to describe the image quality after rain-removal, an evaluation system for rain-removal images from an AI perspective will help us to evaluate the robustness of the rain-removal algorithm and its vulnerability rating. Unlike SSIM and PSNR, Image reconstruction quality evaluation from the perspective of AI depends more on the restoration effect of the potential semantics of the image. This paper analyses the vulnerability of rain-removal algorithms from an AI perspective by quantifying the semantic understanding gap of AI in rain-removal images by taking advantage of the output by potential AI applications on clean background images as an evaluation benchmark. This paper uses Yolo v5 [44] and the Tencent Scene Understanding AI platform [45] as the image reconstruction benchmark. For any AI benchmark, there are:

$$\begin{cases} \mathbb{S} = \dfrac{\sum_i^S \left| \delta_i^{\text{derain}} - \delta_i^{\text{clear}} \right|}{\sum_i^S \delta_i^{\text{derain}}} \\ \mathbb{C} = \dfrac{\sum_i^S \left( c_i^{\text{derain}} - c_i^{\text{clear}} \right) \mathbb{I} \left( \delta_i^{\text{derain}} = \delta_i^{\text{clear}} \right)}{\sum_i^S \mathbb{I}(\delta_i^{\text{derain}} = \delta_i^{\text{clear}})} \end{cases} \quad (14)$$

where $\delta$ represents the one-hot code of an image content given by AI; $S$ is the content of the image as identified by the AI; $\mathbb{I}$ is the indicator function; $\mathbb{S}$ and $\mathbb{C}$ denote the identify error and the confidence bias respectively.

*5.2. Perturbations undetectable to humans*

Table 3 shows the evaluation results on the test dataset of the RainDS. RainCCN can achieve approximately 81% SSIM to clean images in the test set and reach a PSNR value of 27.13. The rain-removal performance is reduced in the case of random spatial variation attacks, with SSIM reduced by 17.3% and PSNR reduced by 16.9%. Under the URA attack, the rain-removal performance is significantly reduced, with SSIM reduced by 39.5% and PSNR reduced by 26.4%.

Table 3. Evaluation results

| Evaluations | Clear Images (CI) | RainCCN | Within URA | Within Random Attack |
|---|---|---|---|---|
| SSIM-Compare with CI | 1 | 0.81 | 0.49 | 0.67 |
| PSNR-Compare with CI | +∞ | 27.13 | 19.96 | 22.53 |
| Identify Error-Tencent AI | 0 | 0.12 | 0.23 | 0.20 |
| Confidence Bias-Tencent AI | 1 | 0.87 | 0.45 | 0.51 |
| Identify Error-Yolo v5 | 0 | 0.05 | 0.09 | 0.08 |
| Confidence Bias-Yolo v5 | 1 | 0.91 | 0.79 | 0.83 |

The rain removal algorithm analyses the association between image pixels to remove the blurring and obscure image scene information by raindrops/rainstreaks. Figure 4 shows the rain-removal and attack results for the ten observed images. These images demonstrate the potential threat to DL-based image rain-removal algorithms:
- Humans cannot distinguish whether a rain-observed sample has been attacked. The perturbations in the "Adv-Rain" images are so similar to the environmental information that human vision cannot analyse them for any potential suspicion.
- Humans are unable to distinguish the presence of malicious attacks or sensor degradation problems from the rain-removal results. Human vision can detect the degradation of rain removal performance from the "Adv-Removal" images due to the perturbation attack, but the cause of the degradation is not directly available. The black-box nature of DL models makes this problem difficult to solve at all.

The presence of these potential threats greatly hinders the use of autonomous driving in high rainfall areas and creates insurmountable difficulties in the investigation of possible accidents. The pixel distribution of the samples further validates the concern that the presence of such threats is linked to technological developments. Figure 5 shows the pixel distribution of a sample in the RGB channels. The results show that the pixel distribution of RainCCN-generated rain-removal images without perturbation is very similar to that of the clear images. However, the URA-generated perturbations minimally modify the observed image pixel distribution and drastically interfere with the pixel restoration capability of RainCCN. The pixel distribution of the rain-removal image shows a different shape from the clear image in the perturbed case. Under the attack of random flow field perturbation, the rain-removal result of the attack reduced the scene information recognition ability of Tencent AI by 66.6% and the confidence of correctly identified scene information by 41.3%; relatively under the attack of URA, the rain-removal result of the attack reduced the scene information recognition ability of Tencent AI by 91.7% and the confidence of correctly identified scene information by 47.2%.

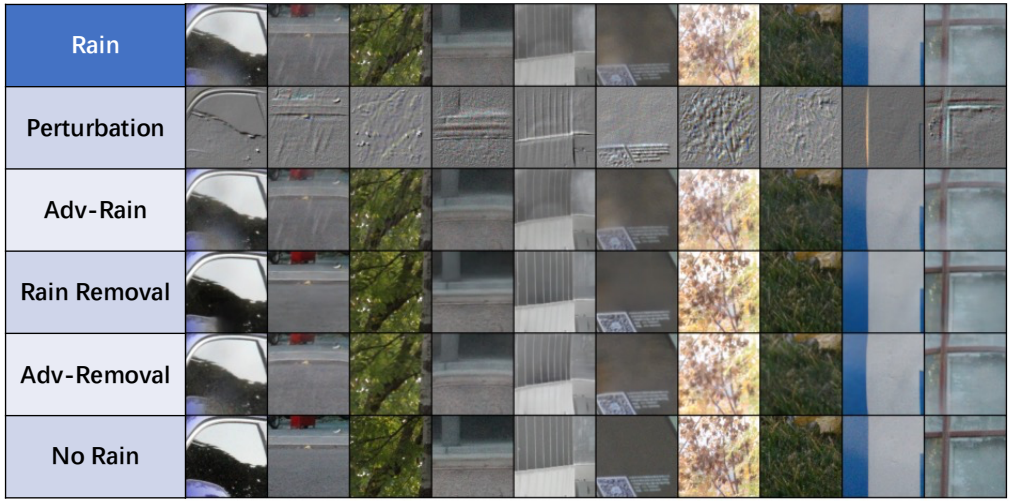

Figure 4. "Rain" shows images observed on a rainy day; "Perturbation" shows the spatial-variational perturbations generated by URA; "Adv-Rain " shows the image of the perturbed observation; "Rain Removal" shows the background image after RainCCN reduction; "Adv-Removal" shows RainCCN attempt to to restore the background image from the scrambled image; 'No Rain' shows a clear background image that is not blurred or obscured by raindrops/rain streaks.

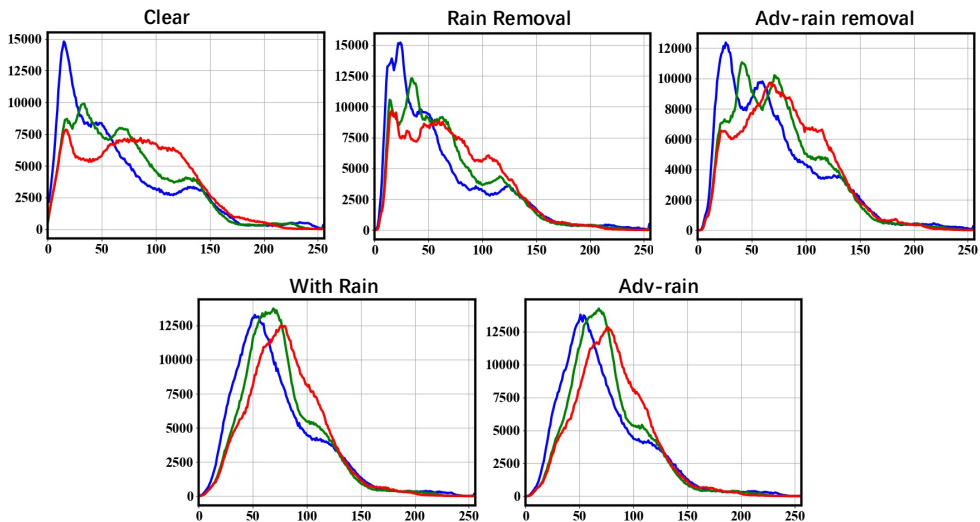

Figure 5. Image pixel analysis: 200.png in the RainDS test set.

## 5.3. AI testing in computer vision

In computer vision, the ability of AI to analyse and understand images is entirely different from humanity, and its ability to analyse and process features at the pixel level has led to a significant evolution in its pattern recognition capabilities compared to people. Table 2 shows the evaluation of potential threats to the DL-based rain-removal algorithm from the perspective of state-of-the-art AI. The rain removal results without the URA attack can be identified by Tencent AI and Yolo v5 with high accuracy and precision. After the attack, the AI of computer vision began to suffer performance degradation. Under the attack of random flow field perturbation, the rain-removal result of the attack reduced the scene information recognition ability of Tencent AI by 66.6% and the confidence of correctly identified scene information by 41.3%; relatively under the attack of URA, the rain-removal result of the attack reduced the scene information recognition ability of Tencent AI by 91.7% and the confidence of correctly identified scene information by 47.2%. The results show that Yolo v5 shows better robustness, but still realises a similar trend.

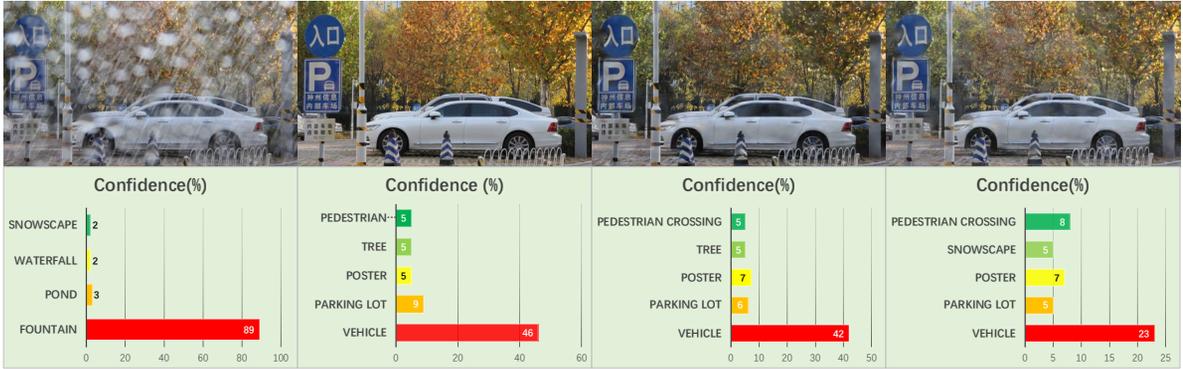

Figure 6. Tencent AI test results. From left to right, it is: rainy day observation image test, clean image test, undisturbed image test and disturbed image test.

Figure 6 shows the scene recognition results of Tencent AI in the sample, where the rainy observation was identified as an artificial fountain with high confidence, which is related to the image being obscured by large rain streaks. Tencent AI gives convincing scene recognition results and corresponding confidence in clean background images. RainCCN was able to provide a satisfactory rain-removal image without the attack that Tencent AI could correctly identify with a reasonable level of confidence. However, the URA attack significantly reduces confidence in correct identification and leads to confusing identification errors. Even though Yolo v5 showed strong robustness, the potential threat was only mitigated rather than resolved. Figure 7 demonstrates the powerful multi-target object recognition capabilities of Yolo v5, which is enhanced by the rain-removal results provided by RainCCN in a rainy environment before it is used for an attack. However, the URA attack still reduces confidence in its correct identification results, especially on objects at the edges of the image. These results demonstrate from an AI perspective that the vulnerability and robustness of image rain-removal algorithms in computer vision-based environmental perception tasks may pose the following threats:

- The robustness of image rain removal algorithms may significantly impact subsequent image intelligence perception tasks. The AI pixel-sensitivity, as demonstrated by the experimental results, implies that a high-performance rain-removal algorithm can effectively improve the AI deployment in rainy weather, but the robustness issues of the rain-removal algorithm may also lead to poor results.
- Potential malicious attacks and sensor degradation may exploit the vulnerability of the image rain-removal algorithm to cause probable traffic accidents. URA, as a universal perturbation generation method, not only has the potential to be deployed and used for criminal purposes in the real world, but can also be seen as a model for sensor degradation. The experimental results demonstrate that the perturbation attack represented by the URA is able to exploit the vulnerability of the rain-removal algorithm to significantly weaken the AI.

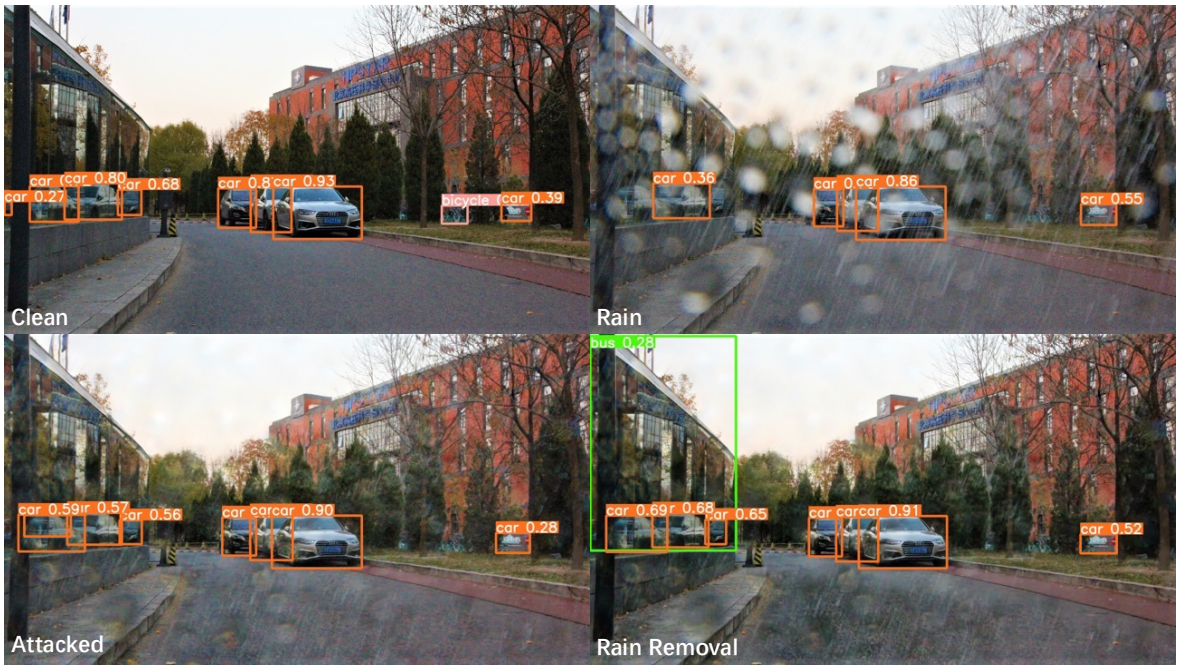

Figure 7. Yolo v5 test results on samples.

## 6. Discussion

This paper validates and explores the possible vulnerability and robustness issues of DL-based rain removal algorithms by constructing a universal image perturbation. The experimental results show that the proposed perturbation generation method, URA, can significantly reduce the image rain-removal performance and exploit this feature to weaken post-attached computer-vision-based AI. It exposes a number of potential threats that computer-vision-based autonomous driving intelligence perception tasks may face:

1. Malicious attacks in real. The proposed perturbation generation method has good potential for practical application and is executable for replication in the real world.
2. Robustness issues of the image rain-removal. Inadequate performance of image rain-removal algorithms and sensor degradation issues may cause AI weakening problems.
3. Difficulties in accident investigation. Undetectable human perturbations in the image rain-removal task of the autonomous driving process can potentially cause serious intelligence perception problems and thus induce safety accidents. Such perturbations are difficult to identify in subsequent accident investigations.

The experimental results also suggest possible mitigation measures:

1. High-performance AI is able to adapt to unclear or perturbed rain-removal images. The Yolo v5 demonstrated resistance to URA in the experimental results.
2. The introduction of physical attributes has the potential to solve the problem of perturbation attacks. The pixel distribution shown in Figure 5 indicates that the distribution of the perturbed observed image in the RGB channel is similar to that of the normal observed image. The luminance, saturation and local structure have some similarities. The physical property-based denoising method is expected to alleviate this problem.

Under the influence of multiple and complex observable factors, this paper has the following limitations:
1. Insufficient targets for the perturbation attack. Due to the limitation of computational resources, the lack of relevant open-source projects and the fact that some of the studies cannot be reproduced, only RainCCN is used as the attack target in this paper. Not only does it achieve SOTA in the current rain-removal problem, but it is also the best result reproduced in this paper. Adding attack targets will be an important task in subsequent research.
2. Limited validation data. There is less high-quality real data in the image rain-removal dataset, and it is difficult to develop and validate perturbation attacks in synthetic data convincingly. Therefore constructing rain-removal datasets from natural driving behaviour in future research would be effective in developing this research methodology.
3. No attack examples in the real world. This paper only shows the perturbation results of the proposed attack method on the dataset and indicates its potential for real-world applications. Due to the lack of relevant hardware devices, this paper cannot show an example of an attack that occurred in the real world. Therefore, the proposed attack method will be further extended, and real-world attack cases and analysis will be given in future research.

## 7. Conclusion

This paper focuses on the potential safety of rain-removal issues in autonomous driving applications. The proposed end-to-end universal perturbation generation method, URA, could reduce the performance of DL-based image rain-removal algorithms and weaken the potential computer-vision-based environment perception AI. Through thorough experiments and results analysis, this paper exposes the potential of DL-based image rain-removal algorithms in autonomous driving applications and provides recommendations for mitigation.